# Giveme5W1H: A Universal System for Extracting Main Events from News Articles

Felix Hamborg[1], Corinna Breitinger[1], Bela Gipp[2]

[1] University of Konstanz, Germany
{firstname.lastname}@uni-konstanz.de
[2] University of Wuppertal, Germany
gipp@uni-wuppertal.de

## ABSTRACT

Event extraction from news articles is a commonly required prerequisite for various tasks, such as article summarization, article clustering, and news aggregation. Due to the lack of universally applicable and publicly available methods tailored to news datasets, many researchers redundantly implement event extraction methods for their own projects. The journalistic 5W1H questions are capable of describing the main event of an article, i.e., by answering *who* did *what*, *when*, *where*, *why*, and *how*. We provide an in-depth description of an improved version of Giveme5W1H, a system that uses syntactic and domain-specific rules to automatically extract the relevant phrases from English news articles to provide answers to these 5W1H questions. Given the answers to these questions, the system determines an article's main event. In an expert evaluation with three assessors and 120 articles, we determined an overall precision of p=0.73, and p=0.82 for extracting the first four W questions, which alone can sufficiently summarize the main event reported on in a news article. We recently made our system publicly available, and it remains the only universal open-source 5W1H extractor capable of being applied to a wide range of use cases in news analysis.

## CCS CONCEPTS

• **Computing methodologies → Information extraction** • Information systems → Content analysis and feature selection • Information systems → Summarization

## KEYWORDS

News Event Detection, 5W1H Extraction, 5W1H Question Answering, Reporter's Questions, Journalist's Questions, 5W QA.

## 1 INTRODUCTION

The extraction of a news article's main event is an automated analysis task at the core of a range of use cases, including news aggregation, clustering of articles reporting on the same event, and news summarization [4, 15]. Beyond computer science, other disciplines also analyze news events, for example, researchers from the social sciences analyze how news outlets report on events in what is known as *frame analyses* [13, 14].

Despite main event extraction being a fundamental task in news analysis, no publicly available method exists that can be applied to the diverse use cases mentioned to capably extract explicit event descriptors from a given article [17]. *Explicit event descriptors* are properties that occur in a text to describe an event, e.g., the phrases in an article that enable a reader to understand what the article is reporting on. The reliable extraction of event-describing phrases also allows later analysis tasks to use common natural language processing (NLP) methods, such as TF-IDF and cosine similarity, including named entity recognition (NER) [10] and named entity disambiguation (NERD) [19] to assess the similarity of two events. State-of-the-art methods for extracting events from articles suffer from three main shortcomings [17]. First, most approaches only detect events implicitly, e.g., by employing topic modeling [2, 42]. Second, they are specialized for the extraction of task-specific properties, e.g., extracting only the number of injured people in an attack [32, 42]. Lastly, some methods extract explicit descriptors, but are not publicly available, or are described in insufficient detail to allow researchers to reimplement the approaches [34, 45, 47, 48].

Last year, we introduced *Giveme5W1H* in the form of a poster abstract [16], which was at that time still an in-progress prototype capable of extracting universally usable phrases that answer the journalistic 5W1H questions, i.e., who did what, when, where, why, and how (see Figure 1). This poster, however, did not disclose or discuss the scoring mechanisms used for determining the best candidate phrases during main event extraction. In this paper, we describe in detail how the improved version of *Giveme5W1H* extracts 5W1H phrases and we describe the results of our evaluation of these improvements. We also introduce an annotated data set, which we created to train our system's model to improve extraction performance. The training data set is available in the online repository (see Section 6) and can be used by other researchers to train their own 5W1H approaches. This paper is relevant to researchers and developers from various disciplines with the shared aim of extracting and analyzing the main events that are being reported on in articles.

**Taliban attacks German consulate in northern Afghan city of Mazar-i-Sharif with truck bomb**

*The death toll from a powerful Taliban truck bombing at the German consulate in Afghanistan's Mazar-i-Sharif city rose to at least six Friday, with more than 100 others wounded in a major militant assault.*

The Taliban said the bombing late Thursday, which tore a massive crater in the road and overturned cars, was a "revenge attack" for US air strikes this month in the volatile province of Kunduz that left 32 civilians dead. [...] The suicide attacker rammed his explosives-laden car into the wall [...].

**Figure 1: News article [1] consisting of title (bold), lead paragraph (italic), and first of remaining paragraphs. Highlighted phrases represent the 5W1H event properties (who did what, when, where, why, and how).**

Our objective is to devise an automated method for extracting the main event being reported on by a given news article. For this purpose, we exclude non-event-reporting articles, such as commentaries or press reviews. First, we define the extracted main event descriptors to be *concise* (requirement *R1*). This means they must be as short as possible and contain only the information describing the event, while also being as long as necessary to contain all information of the event. Second, the descriptors must be of *high accuracy (R2)*. For this reason, we give higher priority to extraction accuracy than execution speed [17]. We also defined that the developed system must achieve a higher extraction accuracy than Giveme5W [17]. Compared to Giveme5W, the system proposed in this paper not only additionally extracts the 'how' answer, but its analysis workflow is more semantics-oriented to address the issues of the previous statistics- and syntax-based extraction. We also publish the first annotated 5W1H dataset, which we use to learn the optimal parameters. In the Giveme5W implementation, the values were based on expert judgement.

The presented system especially benefits: (1) social scientists with limited programming knowledge, who would benefit from ready-to-use main event extraction methods, and (2) computer scientists who are welcome to modify or build on any of the modular components of our system and use our test collection and results as a benchmark for their implementations.

## 2 RELATED WORK

The extraction of 5W1H phrases from news articles is related to closed-domain question answering, which is why some authors call their approaches *5W1H question answering* (QA) systems. Hamborg et al. [17] gave an in-depth overview of 5W1H extraction systems. Thus, we only provide a brief summary of the current state-of-the-art and focus this section on the extraction of the 'how' phrases. Most systems focus only on the extraction of 5W phrases without 'how' phrases (cf. [9, 34, 47, 48]). The authors of prior work do not justify this, but we suspect two reasons.

First, the 'how' question is particularly difficult to extract due to its ambiguity, as we will explain later in this section. Second, 'how' (and 'why') phrases are considered less important in many use cases when compared to the other phrases, particularly those answering the 'who', 'what', 'when', and 'where' (4W) questions

(cf. [21, 40, 49]). For the sake of readability in this section, we will also include approaches that only extract the 5Ws when referring to 5W1H extraction. Aside for the 'how' extraction, the analysis of approaches for 5W1H or 5W-extraction is generally the same.

Systems for 5W1H QA on news texts typically perform three tasks to determine the article's main event [45, 47]: (1) *preprocessing*, (2) *phrase extraction* [10, 25, 36, 47, 48], where for instance linguistic rules are used to extract phrases candidates, and (3) *candidate scoring*, which selects the best answer for each question by employing heuristics, such as the position of a phrase within the document. The input data to QA systems is usually text, such as a full article including the headline, lead paragraph, and main text [36], or a single sentence, e.g., in news ticker format [48]. Other systems use automatic speech recognition (ASR) to convert broadcasts into text [47]. The outcomes of the process are six textual phrases, one for each of the 5W1H questions, which together describe the main event of a given news text, as highlighted in Figure 1. Thus far, no systems have been described in sufficient detail to allow for a reimplementation by other researchers.

Both the 'why' and 'how' question pose a particular challenge in comparison to the other questions. As discussed by Hamborg et al. [17], determining the reason or cause (i.e. 'why') can even be difficult for humans. Often the reason is unknown, or it is only described implicitly, if at all [11]. Extracting the 'how' answer is also difficult, because this question can be answered in many ways. To find 'how' candidates, the system by Sharma et al. extracts the adverb or adverbial phrase within the 'what' phrase [36]. The tokens extracted with this simplistic approach *detail the verb*, e.g., "He drove quickly", but do not answer the *method* how the action was performed (cf. [37]), e.g., by ramming an explosive-laden car into the consulate (in the example in Figure 1), which is a prepositional phrase. Other approaches employ ML [24], but have not been devised for the English language. In summary, few approaches exist that extract 'how' phrases. The reviewed approaches provide no details on their extraction method, and achieve poor results, e.g., they extract adverbs rather than the tool or the method by which an action was performed (cf. [22, 24, 36]).

None of the reviewed approaches output *canonical* or *normalized data*. Canonical output is more concise and also less ambiguous than its original textual form (cf. [46]), e.g., polysemes, such as crane (animal or machine), have multiple meanings. Hence, canonical data is often more useful for subsequent analysis tasks (see Section 1). Phrases containing temporal information or location information may be canonicalized, e.g., by converting the phrases to dates or timespans [7, 38] or to precise geographic positions [29]. Phrases answering the other questions could be canonicalized by employing NERD on the contained NEs, and then linking the NEs to concepts defined in a knowledge graph, such as YAGO [19], or WordNet [31].

While the evaluations of reviewed papers generally indicate sufficient quality to be usable for news event extraction, e.g., the system by Yaman et al. achieved $F_1 = 0.85$ on the Darpa corpus from 2009 [48], the evaluations lack comparability for two reasons. First, no gold standard exists for journalistic 5W1H question answering on news articles. A few datasets exist for automated question answering, specifically for the purpose of disaster tracking [28, 41]; However, these datasets are so specialized to their own use cases that they cannot be applied to the use case of automated journalistic question answering. Another challenge to the evaluation of news event extraction is that the evaluation data sets of previous papers are no longer publicly available [34, 47, 48]. Second, previous papers each used different quality measures, such as precision and recall [9] or error rates [47].

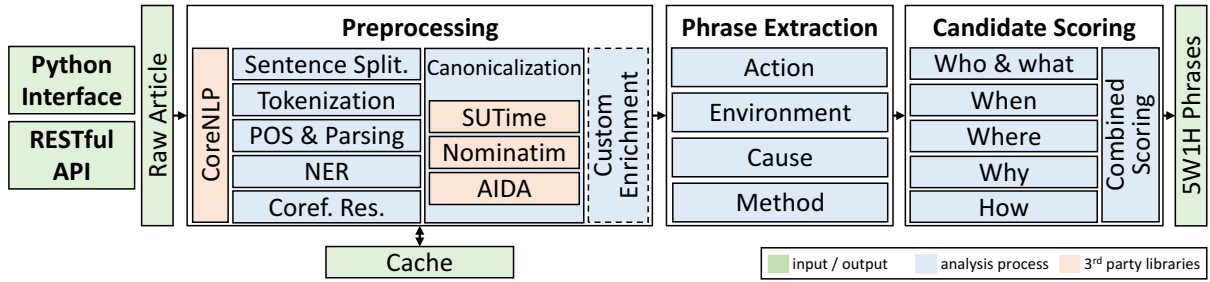

**Figure 2: The three-phases analysis pipeline preprocesses a news text, finds candidate phrases for each of the 5W1H questions, and scores these. Giveme5W1H can easily be accessed via Python and via a RESTful API.**

## 3 GIVEME5W1H: DESCRIPTION OF METHODS AND SYSTEM

*Giveme5W1H* is an open-source main event retrieval system for news articles that addresses the objectives we defined in Section 1. The system extracts 5W1H phrases that describe the most defining characteristics of a news event, i.e., *who* did *what*, *when*, *where*, *why*, and *how*. This section describes the analysis workflow of Giveme5W1H, as shown in Figure 1. Giveme5W1H can be accessed by other software as a Python library and via a RESTful API. Due to its modularity, researchers can efficiently adapt or replace components. For example, researchers can integrate a custom parser or adapt the scoring functions tailored to the characteristics of their data. The system builds on Giveme5W [17], but improves extraction performance by addressing the planned future work directions: Giveme5W1H uses coreference resolution, question-specific semantic distance measures, combined scoring of candidates, and extracts phrases for the 'how' question. The values of the parameters introduced in this section result from a semi-automated search for the optimal configuration of Giveme5W1H using an annotated learning dataset including a manual, qualitative revision (see Section 3.5).

### 3.1 Preprocessing of News Articles

Giveme5W1H accepts as input the full text of a news article, including headline, lead paragraph, and body text. The user can specify these three components as one or separately. Optionally, the article's publishing date can be provided, which helps Giveme5W1H parse relative dates, such as "yesterday at 1 pm".

During *preprocessing*, we use Stanford CoreNLP for sentence splitting, tokenization, lemmatization, POS-tagging, full parsing, NER (with Stanford NER's seven-class model), and pronominal and nominal coreference resolution. Since our main goal is high 5W1H extraction accuracy (rather than fast execution speed), we use the best-performing model for each of the CoreNLP annotators, i.e., the 'neural' model if available. We use the default settings for English in all libraries.

After the initial preprocessing, we bring all NEs in the text into their *canonical* form. Following from requirement R1, canonical information is the preferred output of Giveme5W1H, since it is the most concise form. Because Giveme5W1H uses the canonical information to extract and score 'when' and 'where' candidates, we implement the canonicalization task during preprocessing.

We parse dates written in natural language into canonical dates using SUTime [7]. SUTime looks for NEs of the type date or time and merges adjacent tokens to phrases. SUTime also handles heterogeneous phrases, such as "yesterday at 1 pm", which consist not only of temporal NEs but also other tokens, such as function words. Subsequently, SUTime converts each temporal phrase into a standardized TIMEX3 [44] instance. TIMEX3 defines various types, also including repetitive periods. Since events according to our definition occur at a single point in time, we only retrieve datetimes indicating an exact time, e.g., "yesterday at 6pm", or a duration, e.g., "yesterday", which spans the whole day.

*Geocoding* is the process of parsing places and addresses written in natural language into canonical *geocodes*, i.e., one or more coordinates referring to a point or area on earth. We look for tokens classified as NEs of the type location (cf. [48]). We merge adjacent tokens of the same NE type within the same sentence constituent, e.g., within the same NP or VP. Similar to temporal phrases, locality phrases are often heterogeneous, i.e., they do not only contain temporal NEs but also function words. Hence, we introduce a locality phrase merge range $r_{where} = 1$, to merge phrases where up to $r_{where}$ arbitrary NE tokens are allowed between two location NEs. Lastly, we geocode the merged phrases with Nominatim[1], which uses free data from OpenStreetMap.

We canonicalize NEs of the remaining types, e.g., persons and organizations, by linking NEs to concepts in the YAGO graph [30] using AIDA [19]. The YAGO graph is a state-of-the-art knowledge base, where nodes in the graph represent semantic concepts that are connected to other nodes through attributes and relations. The data is derived from other well-established knowledge bases, such as Wikipedia, WordNet, WikiData, and GeoNames [39].

---

[1] https://github.com/openstreetmap/Nominatim, v3.0.0

## 3.2 Phrase Extraction

Giveme5W1H performs four independent extraction chains to retrieve the article's main event: (1) the action chain extracts phrases for the 'who' and 'what' questions, (2) environment for 'when' and 'where', (3) cause for 'why', and (4) method for 'how'.

The *action extractor* identifies who did what in the article's main event. The main idea for retrieving 'who' candidates is to collect the subject of each sentence in the news article. Therefore, we extract the first NP that is a direct child to the sentence in the parse tree, and that has a VP as its next right sibling (cf. [5]). We discard all NPs that contain a child VP, since such NPs yield lengthy 'who' phrases. Take, for instance, this sentence: "*((NP) Mr. Trump, ((VP) who stormed to a shock election victory on Wednesday)), ((VP) said it was [...])*", where "who stormed [...]" is the child VP of the NP. We then put the NPs into the list of 'who' candidates. For each 'who' candidate, we take the VP that is the next right sibling as the corresponding 'what' candidate (cf. [5]). To avoid long 'what' phrases, we cut VPs after their first child NP, which long VPs usually contain. However, we do not cut the 'what' candidate if the VP contains at most $l_{\text{what,min}} = 3$ tokens, and the right sibling to the VP's child NP is a prepositional phrase (PP). This way, we avoid short, undescriptive 'what' phrases. For instance, in the simplified example: "*((NP) The microchip) ((VP) is ((NP) part) ((PP) of a wider range of the company's products)).*", the truncated VP "is part" contains no descriptive information; Hence, our presented rules prevent this truncation.

The *environment extractor* retrieves phrases describing the temporal and locality context of the event. To determine 'when' candidates, we take TIMEX3 instances from preprocessing. Similarly, we take the geocodes as 'where' candidates.

The *cause extractor* looks for linguistic features indicating a causal relation within a sentence's constituents. We look for three types of cause-effect indicators (cf. [25, 26]): *causal conjunctions*, *causative adverbs*, and *causative verbs*. Causal conjunctions, e.g. "due to", "result of", and "effect of", connect two clauses, whereas the second clause yields the 'why' candidate. For causative adverbs, e.g., "therefore", "hence", and "thus", the first clause yields the 'why' candidate. If we find that one or more subsequent tokens of a sentence match with one of the tokens adapted from Khoo et al. [25], we take all tokens on the right (causal conjunction) or left side (causative adverb) as the 'why' candidate.

Causative verbs, e.g. "activate" and "implicate", are contained in the middle VP of the causative NP-VP-NP pattern, whereas the last NP yields the 'why' candidate [11, 26]. For each NP-VP-NP pattern we find in the parse-tree, we determine whether the VP is causative. To do this, we extract the VP's verb, retrieve the verb's synonyms from WordNet [31] and compare the verb and its synonyms with the list of causative verbs from Girju [11], which also extended by their synonyms (cf. [11]). If there is at least one match, we take the last NP of the causative pattern as the 'why' candidate. To reduce false positives, we check the NP and VP for the causal constraints for verbs proposed by Girju [11].

The *method extractor* retrieves 'how' phrases, i.e., the method by which an action was performed. The combined method consists of two subtasks, one analyzing *copulative conjunctions*, the other looking for *adjectives* and *adverbs*. Often, sentences with a copulative conjunction contain a method phrase in the clause that follows the copulative conjunction, e.g., "after [the train came off the tracks]". Therefore, we look for copulative conjunctions compiled from [33]. If a token matches, we take the right clause as the 'how' candidate. To avoid long phrases, we cut off phrases longer than $l_{\text{how,max}} = 10$ tokens. The second subtask extracts phrases that consist purely of adjectives or adverbs (cf. [36]), since these often represent how an action was performed. We use this extraction method as a fallback, since we found the copulative conjunction-based extraction too restrictive in many cases.

## 3.3 Candidate Scoring

The last task is to determine the best candidate of each 5W1H. The scoring consists of two sub-tasks. First, we score candidates independently for each of the 5W1H questions. Second, we perform a combined scoring where we adjust scores of candidates of one question dependent on properties, e.g., position, of candidates of other questions. For each question $q$, we use a scoring function that is composed as a weighted sum of $n$ scoring factors: $s_q = \sum_{i=0}^{n-1} w_{q,i} s_{q,i}$, where $w_{q,i}$ is the weight of the scoring factor $s_{q,i}$.

To score 'who' candidates, we define three scoring factors: the candidate shall occur in the article (1) *early* and (2) *often*, and (3) contain a *named entity*. The first scoring factor targets the concept of the *inverse pyramid* [8]: news mention the most important information, i.e., the main event, early in the article, e.g., in the headline and lead paragraph, while later paragraphs contain details. However, journalists often use so called hooks to get the reader's attention without revealing all content of the article [35]. Hence, for each candidate, we also consider the frequency of similar phrases in the article, since the primary actor involved in the main event is likely to be mentioned frequently in the article. Furthermore, if a candidate contains a NE, we will score it higher, since in news, the actors involved in events are often NEs, e.g., politicians. Table 1 shows the weights and scoring factors.

**Table 1: Weights and scoring factors for 'who' phrases**

| $i$ | $w_{\text{who},i}$ | $s_{\text{who},i}$ |
|---|---|---|
| 0 (position) | .9 | $\text{pos}(c) = 1 - \dfrac{n_{\text{pos}}(c)}{d_{\text{len}}}$ |
| 1 (frequency) | .095 | $\text{f}(c) = \dfrac{n_f(c)}{\max_{c' \in C}\left(n_f(c')\right)}$ |
| 2 (type) | .005 | $\text{NE}(c)$ |

$n_{\text{pos}}(c)$ is the position measured in sentences of candidate $c$ within the document, $n_f(c)$ the frequency of phrases similar to $c$ in the document, and $\text{NE}(c) = 1$ if $c$ contains an NE, else 0 (cf. [10]). To measure $n_f(c)$ of the actor in candidate $c$, we use the number of the actor's coreferences, which we extracted during coreference resolution (see Section 3.1). This allows Giveme5W1H to recognize and count name variations, as well as pronouns. Due to the strong relation between agent and action, we rank VPs according to their NPs' scores. Hence, the most likely VP is the sibling in the parse tree of the most likely NP: $s_{\text{what}} = s_{\text{who}}$.

We score temporal candidates according to four scoring factors: the candidate shall occur in the article (1) *early* and (2) *often*. It should also be (3) *close to the publishing date* of the article, and (4) of a relatively *short duration*. The first two scoring factors have the same motivation as in the scoring of 'who' candidates. The idea for the third scoring factor, close to the publishing date, is that events reported on by news articles often occurred on the same day or on the day before the article was published. For example, if a candidate represents a date one or more years in the past before the publishing date of the article, the candidate will achieve the lowest possible score in the third scoring factor. The fourth scoring factor prefers temporal candidates that have a short duration, since events according to our definition happen during a specific point in time with a short duration. We logarithmically normalize the duration factor between one minute and one month (cf. [49]). The resulting scoring formula for a temporal candidate $c$ is the sum of the weighted scoring factors shown in Table 2.

**Table 2: Weights and scoring factors for 'when' phrases**

| $i$ | $w_{\text{when},i}$ | $s_{\text{when},i}$ |
|---|---|---|
| 0 (position) | .24 | $\text{pos}(c)$ |
| 1 (frequency) | .16 | $\text{f}(c)$ |
| 2 (closeness) | .4 | $1 - \min 1, \dfrac{\Delta_s(c, d_{\text{pub}})}{\text{e}_{\max}}$ |
| 3 (duration) | .2 | $1 - \min 1, \dfrac{\log s(c) - \log s_{\min}}{\log s_{\max} - \log s_{\min}}$ |

To count $n_f(c)$, we determine two TIMEX3 instances as similar if their start and end-dates are at most 24h apart. $\Delta_s(c, d_{\text{pub}})$ is the difference in seconds of candidate $c$ and the publication date of the news article $d_{\text{pub}}$, $s(c)$ the duration in seconds of $c$, and the normalization constants $\text{e}_{\max} \approx 2.5\text{Ms}$ (one month in seconds), $s_{\min} = 60\text{s}$, and $s_{\max} \approx 31\text{Ms}$ (one year).

The scoring of *location* candidates follows four scoring factors: the candidate shall occur (1) *early* and (2) *often* in the article. It should also be (3) *often geographically contained* in other location candidates and be (4) *specific*. The first two scoring factors have the same motivation as in the scoring of 'who' and 'when' candidates. The second and third scoring factors aim to (1) find locations that occur often, either by being similar to others, or (2) by being contained in other location candidates. The fourth scoring factor favors specific locations, e.g., Berlin, over broader mentions of location, e.g., Germany or Europe. We logarithmically normalize the location specificity between $a_{\min} = 225m^2$ (a small property's size) and $a_{\max} = 530,000km^2$ (approx. the mean area of all countries [43]). We discuss other scoring options in Section 5. The used weights and scoring factors are shown in Table 3. We measure $n_f(c)$, the number of similar mentions of candidate $c$, by counting how many other candidates have the same Nominatim place ID. We measure $n_e(c)$ by counting how many other candidates are geographically contained within the bounding box of $c$, where $a(c)$ is the area of the bounding box of $c$ in square meters.

**Table 3: Weights and scoring factors for 'where' phrases**

| $i$ | $w_{\text{where},i}$ | $s_{\text{where},i}$ |
|---|---|---|
| 0 (position) | .37 | $\text{pos}(c)$ |
| 1 (frequency) | .3 | $\text{f}(c)$ |
| 2 (containment) | .3 | $\dfrac{n_e(c)}{\max_{c' \in C}(n_e(c'))}$ |
| 3 (specificity) | .03 | $1 - \min 1, \dfrac{\log a(c) - \log a_{\min}}{\log a_{\max} - \log a_{\min}}$ |

Scoring *causal* candidates was challenging, since it often requires semantic interpretation of the text and simple heuristics may fail [11]. We define two objectives: candidates shall (1) occur *early* in the document, and (2) their *causal type* shall be reliable [26]. The second scoring factor rewards causal types with low ambiguity (cf. [3, 11]), e.g., "because" has a very high likelihood that the subsequent phrase contains a cause [11]. The weighted scoring factors are shown in Table 4. The causal type $\text{TC}(c) = 1$ if $c$ is extracted due to a causal conjunction, 0.62 if it starts with a causative RB, and 0.06 if it contains a causative VB (cf. [25, 26]).

**Table 4: Weights and scoring factors for 'why' phrases**

| $i$ | $w_{\text{why},i}$ | $s_{\text{why},i}$ |
|---|---|---|
| 0 (position) | .56 | $\text{pos}(c)$ |
| 1 (type) | .44 | $\text{CT}(c)$ |

The scoring of *method* candidates uses three simple scoring factors: the candidate shall occur (1) *early* and (2) *often* in the news article, and (3) their *method type* shall be reliable. The weighted scoring factors for method candidates are shown in Table 5.

**Table 5: Weights and scoring factors for 'how' phrases**

| $i$ | $w_{\text{how},i}$ | $s_{\text{how},i}$ |
|---|---|---|
| 0 (position) | .23 | $\text{pos}(c)$ |
| 1 (frequency) | .14 | $\text{f}(c)$ |
| 2 (type) | .63 | $\text{TM}(c)$ |

The method type $\text{TM}(c) = 1$ if $c$ is extracted because of a copulative conjunction, else 0.41. We determine the number of mentions of a method phrase $n_f(c)$ by the term frequency (including inflected forms) of its most frequent token (cf. [45]).

The final sub-task in candidate scoring is *combined scoring*, which adjusts scores of candidates of a single 5W1H question depending on the candidates of other questions. To improve the scoring of method candidates, we devise a combined sentence-distance scorer. The assumption is that the method of performing an action should be close to the mention of the action. The resulting equation for a method candidate $c$ given an action candidate $a$ is:

$$s_{\text{how,new}}(c, a) = s_{\text{how}}(c) - w_0 \frac{|n_{\text{pos}}(c) - n_{\text{pos}}(a)|}{d_{\text{len}}} \quad (1)$$

where $w_0 = 1$. Section 5 describes additional scoring approaches.

## 3.4 Output

The highlighted phrases in Figure 1 are candidates extracted by Giveme5W1H for each of the 5W1H event properties of the shown article. Giveme5W1H enriches the returned phrases with additional information that the system extracted for its own analysis or during *custom enrichment*, with which users can integrate their own preprocessing. The additional information for each token is its POS-tag, parse-tree context, and NE type if applicable. Enriching the tokens with this information increases the efficiency of the overall analysis workflow in which Giveme5W1H may be embedded, since later analysis tasks can reuse the information.

For the temporal phrases and locality phrases, Giveme5W1H also provides their canonical forms, i.e., TIMEX3 instances and geocodes. For the news article shown in Figure 1, the canonical form of the 'when' phrase represents the entire day of November 10, 2016. The canonical geocode for the 'where' phrase represents the coordinates of the center of the city Mazar-i-Sharif (36°42'30.8"N 67°07'09.7"E), where the bounding box represents the area of the city, and further information from OSM, such as a canonical name and place ID, which uniquely identifies the place. Lastly, Giveme5W1H provides linked YAGO concepts [30] for other NEs.

## 3.5 Parameter Learning

Determining the best values for the parameters introduced in Section 3, e.g., weights of scoring factors, is a supervised ML problem [6]. Since there is no gold standard for journalistic 5W1H extraction on news (see Section 2), we created an annotated dataset.

The dataset is available in the open-source repository (see Section 6). To facilitate diversity in both content and writing style, we selected 13 major news outlets from the U.S. and the UK. We sampled 100 articles from the news categories politics, disaster, entertainment, business and sports for November $6^{th} - 14^{th}$, 2016. We crawled the articles [18] and manually revised the extracted information to ensure that it was free of extraction errors.

We asked three annotators (graduate IT students, aged between 22 and 26) to read each of the 100 news articles and to annotate the single most suitable phrase for each 5W1H question. Finally, for each article and question, we combined the annotations using a set of combination rules, e.g., if all phrases were semantically equal, we selected the most concise phrase, or if there was no agreement between the annotators, we selected each annotator's first phrase, resulting in three semantically diverging but valid phrases. We also manually added a TIMEX3 instance to each 'when' annotation, which was used by the error function for 'when'. The intercoder reliability was $ICR_{ann} = 0.81$, measured using average pairwise percentage agreement.

We divided the dataset into two subsets for training (80% randomly sampled articles) and testing (20%). To find the optimal parameter values for our extraction method, we first computed for each parameter configuration the mean error (ME) on the training set. To measure the ME of a configuration, we devised three error functions to measure the semantic distance between candidate phrases and annotated phrases. For the textual candidates, i.e., who, what, why, and how, we used the Word Mover's Distance

(WMD) [27]. WMD is a state-of-the-art generic measure for semantic similarity of two phrases. For 'when' candidates, we computed the difference in seconds between candidate and annotation. For 'where' candidates, we computed the distance in meters between both coordinates. We linearly normalized all measures.

We then validated the 5% best performing configurations on the test set and discarded all configurations that yielded a significantly different ME. Finally, we selected the best performing parameter configuration for each question.

## 4 EVALUATION

We use the same evaluation rules and procedure as described by Hamborg et al. [17] but employed a larger dataset of 120 news articles, which we sampled from the BBC dataset [12] in order to conduct a survey with three assessors. The dataset contains 24 news articles in each of the following categories: business (*Bus*), entertainment (*Ent*), politics (*Pol*), sport (*Spo*), and tech (*Tec*)). We asked the assessors to read one article at a time. After reading each article, we showed the assessors the 5W1H phrases that had been extracted by the system and asked them to judge the relevance of each answer on a 3-point scale: *non-relevant* (if an answer contained no relevant information, score $s = 0$), *partially relevant* (if only part of the answer was relevant or if information was missing, $s = 0.5$), and *relevant* (if the answer was completely relevant without missing information, $s = 1$).

Table 6 shows the *mean average generalized precision* (MAgP), a score suitable for multi-graded relevance assessments [17, 23]. MAgP was 0.73 over all categories and questions. If only considering the first 4Ws, which the literature considers as sufficient to represent an event (cf. [21, 40, 49]), overall MAgP was 0.82.

**Table 6: ICR and MAgP-Performance of Giveme5W1H**

| Question | *ICR* | Bus | Ent | Pol | Spo | Tec | Avg. |
|---|---|---|---|---|---|---|---|
| Who | .93 | .98 | .88 | .89 | .97 | .90 | .92 |
| What | .88 | .85 | .69 | .89 | .84 | .66 | .79 |
| When | .89 | .55 | .91 | .79 | .81 | .82 | .78 |
| Where | .95 | .82 | .63 | .85 | .79 | .80 | .78 |
| Why | .96 | .48 | .62 | .42 | .45 | .42 | .48 |
| How | .87 | .63 | .58 | .68 | .51 | .65 | .61 |
| Avg. all | *.91* | .72 | .72 | .75 | .73 | .71 | **.73** |
| Avg. 4W | *.91* | .80 | .78 | .86 | .85 | .80 | **.82** |

Of the few existing approaches capable of extracting phrases that answer all six 5W1H questions (see Section 2), only one publication reported the results of an evaluation: the approach developed by Khodra achieved a precision of 0.74 on Indonesian articles [24]. Others did not conduct any evaluation [36] or only evaluated the extracted 'who' and 'what' phrases of Japanese news articles [22].

We also investigated the performance of systems that are only capable of extracting 5W phrases. Our system achieves a $MAgP_{5W} = 0.75$, which is 0.05 higher than the MAgP of Giveme5W [17]. We also compared the performance with other systems, despite the difficulties mentioned by Hamborg et al. [17]: other systems were tested on non-disclosed datasets [34, 47, 48], they were translated from other languages [34], they were devised for different languages [22, 24, 45], or they used different evaluation measures, such as error rates [47] or binary relevance assessments [48], which are both not optimal because of the non-binary relevance of 5W1H answers (cf. [23]). Finally, none of the related systems have been made publicly available or have been described in sufficient detail to enable a re-implementation, which was the primary motivation for our research (see Section 1).

Therefore, a direct comparison of the results and related work was not possible. Compared to the fraction of correct 5W answers by the best system by Parton et al. [34], Giveme5W1H achieves a 0.12 higher $MAgP_{5W}$. The best system by Yaman et al. achieved a precision $P_{5W} = 0.89$ [48], which is 0.14 higher than our $MAgP_{5W}$ and – as a rough approximation of the best achievable precision [20] – surprisingly almost identical to the ICR of our assessors.

We found that different forms of journalistic presentation in the five news categories of the dataset led to different extraction performance. Politics articles, which yielded the best performance, mostly reported on single events. The performance on sports articles was unexpectedly high, even though they not only report on single events but also are background reports or announcements, for which event detection is more difficult. Determining the 'how' in sports articles was difficult ($MAgP_{how} = 0.51$), since often articles implicitly described the method of an event, e.g., how one team won a match, by reporting on multiple key events during the match. Some categories, such as entertainment and tech, achieved lower extraction performances, mainly because they often contained much background information on earlier events and the actors involved.

## 5  DISCUSSION AND FUTURE WORK

Most importantly, we plan to improve the extraction quality of the 'what' question, being one of the important 4W questions. We aim to achieve an extraction performance similar to the performance of the 'who' extraction ($MAgP_{who} = 0.91$), since both are strongly related. In our evaluation, we identified two main issues: (1) joint extraction of optimal 'who' candidates with non-optimal 'what' candidates and (2) cut-off 'what' candidates. In some cases (1), the headline contained a concise 'who' phrase but the 'what' phrase did not contain all information, e.g., because it only aimed to catch the reader's interest, a journalistic hook (Section 2). We plan to devise separate extraction methods for both questions. Thereby, we need to ensure that the top candidates of both questions fit to each other, e.g., by verifying that the semantic concept of the answer of each question, e.g., represented by the nouns in the 'who' phrase, or verbs in the 'what' phrase, co-occur in at least one sentence of the article. In other cases (2), our strategy to avoid too detailed 'what' candidates (Section 3.2) cut off the relevant information, e.g., "widespread corruption in the finance ministry has cost it $2m", in which the underlined text was cut off. We will investigate dependency parsing and further syntax rules, e.g., to always include the direct object of a transitive verb.

For 'when' and 'where' questions, we found that in some cases an article does not explicitly mention the main event's date or lo-cation. The date of an event may be implicitly defined by the reported event, e.g., "in the final of the Canberra Classic". The location may be implicitly defined by the main actor, e.g., "Apple Postpones Release of [...]", which likely happened at the Apple headquarters in Cupertino. Similarly, the proper noun "Stanford University" also defines a location. We plan to investigate how we can use the YAGO concepts, which are linked to NEs, to gather further information regarding the date and location of the main event. If no date can be identified, the publishing date of the article or the day before it might sometimes be a suitable fallback date.

Using the standardized TIMEX3 instances from SUTime is an improvement ($MAgP_{when} = 0.78$) over a first version, where we used dates without a duration ($MAgP_{when} = 0.72$).

The extraction of 'why' and 'how' phrases was most challenging, which manifests in lower extraction performances compared to the other questions. One reason is that articles often do not explicitly state a single cause or method of an event, but implicitly describe this throughout the article, particularly in sports articles (see Section 5). In such cases, NLP methods are currently not advanced enough to find and abstract or summarize the cause or method (see Section 3.3). However, we plan to improve the extraction accuracy by preventing the system from returning false positives. For instance, in cases where no cause or method could be determined, we plan to introduce a score threshold to prevent the system from outputting candidates with a low score, which are presumably wrong. Currently, the system always outputs a candidate if at least one cause or method was found.

To improve the performance of all textual questions, i.e., who, what, why, and how, we will investigate two approaches. First, we want to improve measuring a candidate's frequency, an important scoring factor in multiple questions (Section 3.3). We currently use the number of coreferences, which does not include synonymous mentions. We plan to count the number of YAGO concepts that are semantically related to the current candidate. Second, we found that a few top candidates of the four textual questions were semantically correct but only contained a pronoun referring to the more meaningful noun. We plan to add the coreference's original mention to extracted answers.

## 6  CONCLUSION

In this paper, we proposed *Giveme5W1H*, the first open-source system that extracts answers to the journalistic 5W1H questions, i.e., *who* did *what*, *when*, *where*, *why*, and *how*, to describe a news article's main event. The system canonicalizes temporal mentions in the text to standardized TIMEX3 instances, locations to geocoordinates, and other NEs, e.g., persons and organizations, to unique concepts in a knowledge graph. The system uses syntactic and domain-specific rules to extract and score phrases for each 5W1H question. Giveme5W1H achieved a mean average generalized precision (MAgP) of 0.73 on all questions, and an MAgP of 0.82 on the first four W questions (who, what, when, and where), which alone can represent an event. Extracting the answers to 'why' and 'how' performed more poorly, since articles often only imply causes and methods. Answering the 5W1H questions is at the core of understanding any article, and thus an essential task in many research efforts that analyze articles. We hope that redundant implementations and non-reproducible evaluations can be avoided with Giveme5W1H as the first universally applicable, modular, and open-source 5W1H extraction system. In addition to benefiting developers and computer scientists, our system especially benefits researchers from the social sciences, for whom automated 5W1H extraction was previously not made accessible.

Giveme5W1H and the datasets for training and evaluation are available at: **https://github.com/fhamborg/Giveme5W1H**